\newif\ifpeerreview
\peerreviewfalse  

\ifpeerreview
  \documentclass[peerreviewca]{IEEEtran}
\else
  \documentclass[conference]{IEEEtran}
\fi

\IEEEoverridecommandlockouts
\usepackage{booktabs}
\usepackage{cite}
\usepackage{amsmath,amssymb,amsfonts}
\usepackage{graphicx}
\usepackage{textcomp}
\def\BibTeX{{\rm B\kern-.05em{\sc i\kern-.025em b}\kern-.08em
    T\kern-.1667em\lower.7ex\hbox{E}\kern-.125emX}}

\usepackage{algorithm}
\usepackage{tabularx}
\usepackage{algpseudocode}
\usepackage{xcolor}
\usepackage{float}

\newcommand{\ourAlgorithm}{\emph{MechDetect}}

\newcommand{\R}{\mathbb{R}}

\usepackage[hidelinks]{hyperref}

\begin{document}
\def\sectionautorefname{Section}
\def\subsectionautorefname{Section}%

\title{\ourAlgorithm{}: Detecting Data-Dependent Errors
}

\author{
\IEEEaftertitletext{\vspace{-0.5cm}}

\IEEEauthorblockN{Philipp Jung* \thanks{* These authors contributed equally to this research.}}
\IEEEauthorblockA{\textit{Berlin University of Applied}\\
\textit{Sciences and Technology}\\
Berlin, Germany \\
phillip.jung@bht-berlin.de}
\and
\IEEEauthorblockN{Nicholas Chandler*}
\IEEEauthorblockA{\textit{Berlin University of Applied}\\
\textit{Sciences and Technology}\\
Berlin, Germany \\
nich1834@bht-berlin.de}
\and
\IEEEauthorblockN{Sebastian Jäger}
\IEEEauthorblockA{\textit{Berlin University of Applied}\\
\textit{Sciences and Technology}\\
Berlin, Germany \\
sebastian.jaeger@bht-berlin.de}
\and
\IEEEauthorblockN{Felix Biessmann}
\IEEEauthorblockA{\textit{Berlin University of Applied}\\
\textit{Sciences and Technology,}\\
\textit{Einstein Center Digital Future}\\
Berlin, Germany \\
felix.biessmann@bht-berlin.de}
}

\maketitle

\ifpeerreview
    \IEEEpeerreviewmaketitle
\fi

\begin{abstract}
Data quality monitoring is a core challenge in modern information processing systems. While many approaches to detect data errors or shifts have been proposed, few studies investigate the mechanisms governing error generation. We argue that knowing how errors were generated can be key to tracing and fixing them. In this study, we build on existing work in the statistics literature on missing values and propose \ourAlgorithm{}, a simple algorithm to investigate error generation mechanisms. Given a tabular data set and a corresponding error mask, the algorithm estimates whether or not the errors depend on the data using machine learning models. Our work extends established approaches to detect mechanisms underlying missing values and can be readily applied to other error types, provided that an error mask is available. We demonstrate the effectiveness of \ourAlgorithm{} in experiments on established benchmark datasets.
\end{abstract}


\section{Introduction}
\label{sec:introduction}
Handling errors in tabular data is a long-standing problem that affects both scientific research and industry applications~\cite{schelter_challenges_2015,biessmannAutomatedDataValidation}.
Given the importance of data quality in modern information processing systems, there are surprisingly few studies on the mechanisms governing the generation of errors in tabular data sets. One exception is the work on missing data in the statistics literature~\cite{rubinInferenceMissingData1976,endersAppliedMissingData2010,vanBuuren2018}. 
Missingness patterns have been formalized to capture the dependency of missing data on tabular data~\cite{rubinInferenceMissingData1976}. To investigate these dependencies, researchers employ visualizations~\cite{Templ2012Exploring} or statistical tests~\cite{little1988test}.
Identifying the missingness mechanism, or in general error generating mechanisms, remains challenging, particularly in high-dimensional data.
Knowledge of the error generating mechanism can be helpful in research and applications, as different mechanisms can be remedied with appropriate, though sometimes different, cleaning procedures~\cite{jagerBenchmarkDataImputation2021}.
 
Here, we propose \ourAlgorithm{}, a novel data-driven approach for inferring how the error generating mechanism depends on the underlying data distribution.
To validate the effectiveness of \ourAlgorithm{}, we perform an empirical analysis on $101$ datasets from existing machine learning and data cleaning benchmarks~\cite{grinsztajn2022tree,fischer2023openmlctr, bischl2021openmlbenchmarkingsuites, jagerDataImputationData2024}.

\begin{figure*}
  \includegraphics[width=\textwidth]{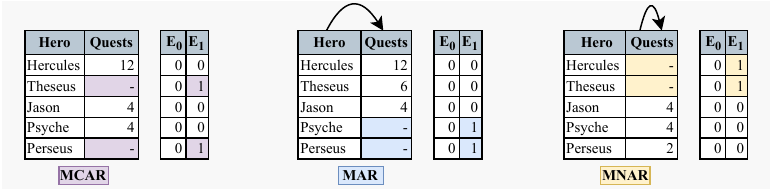}
  \caption{Illustration of error generation mechanisms as formalized for missing values in the statistics literature. 
  Arrows between columns denote a statistical dependency, \textbf{E\textsubscript{0}} and \textbf{E\textsubscript{1}} are the columns of $E$, the binary error mask of the respective table.
  Errors in the left table are independent from data, or Missing Completely At Random (MCAR).
  In the Missing At Random (MAR) example, \textbf{Quests} are missing if the corresponding \textbf{Hero}'s name start with the letter P. Thus, there is a dependency between the error mask and the column \textbf{Hero}, which determines the missing values' positions.
  Finally, in the Missing Not At Random (MNAR) case, values in \textbf{Quests} are missing if they are greater than 5, meaning that $E$ depends on values in \textbf{Quests}.}
  \label{fig:teaser}
\end{figure*}

\section{Related Work}
\label{sec:related-work}
Error generating mechanisms have been studied primarily in the literature on missing values established by Rubin~\cite{rubinInferenceMissingData1976}, who defined conditions under which researchers can safely ignore the mechanism causing missing data. For a detailed overview of the topic of missing data and remediation techniques, we refer to van Buuren~\cite{endersAppliedMissingData2010}. The general idea is to categorize errors---in this case, missing values---into subcategories: those that are independent from data, referred to as missing completely at random (MCAR), and those that depend on data, referred to as missing at random (MAR) or missing not at random (MNAR). The differences are illustrated in \autoref{fig:teaser}. An approach to the categorization and detection of missingness patterns is visualization as in Templ et al.~\cite{Templ2012Exploring}. However, this requires manual supervision and is not inherently quantitative. In addition, various statistical testing methodologies for detecting missingness patterns have been developed~\cite{little1988test, li2015nonparametric, spohn2025pklm, wangScoreTestMissing2023}. 

Little’s MCAR test~\cite{little1988test} utilizes a parametric approach based on likelihood-ratio statistics and assumptions of multivariate normality. These assumptions are often violated in practice which motivated further development. Li and Yu~\cite{li2015nonparametric} relax the distributional assumptions of previous work and propose a nonparametric test that compares distributions of the data across different missing-pattern groups, still focusing on testing for MCAR missingness. More recently, Spohn et al.~\cite{spohn2025pklm} introduced a nonparametric MCAR test that leverages random forest models and Kullback-Leibler Divergence to compare missingness distributions projected onto the variable space. These tests 
do not account for MAR or MNAR missing values, which are both characterized by a dependency of the error distribution on the distribution of observed (MAR) or unobserved (MNAR) values in a table. 
Wang~et~al.~\cite{wangScoreTestMissing2023} introduce a score test to discriminate between MNAR and MAR missing values. Their approach relies on a linear model which limits the capability of the approach to capture non-linear dependency mechanisms between data and errors. Finally, an application of an algorithm similar to \ourAlgorithm{} is given in Dekermanjian~et~al.~\cite{Dekermanjian2022}. The authors use a random forest model to directly classify missingness as either MCAR or MNAR before applying an appropriate data imputation technique. This method still neglects the case of differentiating between MNAR and MAR, a case \ourAlgorithm{} addresses.
\ourAlgorithm{} builds upon the previously mentioned approaches, requires different assumptions, and learns non-linear dependencies between errors and data. It also addresses the three MCAR, MAR, and MNAR distributions.

\section{Methods}
\label{sec:methods}
We design \ourAlgorithm{} to determine the error mechanism of column $j$ by taking the data $X$ and the error mask $E_{:,j}$ as input. This is shown in pseudocode in Algorithm~\ref{alg:mechdetect}.
For this, we construct three different machine learning tasks, train models and compare their performances using non-parametric statistical tests.
An example for this two-step decision process is depicted in~\autoref{fig:testing-tree}.
An assumption \ourAlgorithm{} makes is that the error mask is available when the procedure is run.
This can often be obtained from the data. For instance, in the case of missing values, the error mask is simply the binary mask of the locations of the missing values. This is the approach studied in the experiments below. Error masks for other types of errors can in practice be obtained with outlier detection~\cite{pit--claudelOutlierDetectionHeterogeneous,jager24a} or other approaches~\cite{mahdaviRahaConfigurationFreeError2019,heidariHoloDetectFewShotLearning2019}. Metadata such as constraints~\cite{xuchuHolisticDataCleaning2013} can also be used for creating an error mask.
A subset of the correct data $X$ can be derived through manual or constraint-based correction~\cite{rekatsinasHoloCleanHolisticData2017,mahdaviBaranEffectiveError2020}.

\subsection{Notation}
Consider a matrix $X \in \R^{N \times D}$ representing tabular data, with $N$ rows/observations and $D$ columns/variables. The $i$th row of the table is $X_{i,:} \in \R^{D}$, the $j$th column of the table is $X_{:,j} \in \R^{N}$, and the data excluding column $j$ is $X_{:,-j} \in \R^{N \times (D-1)}$.
The matrix $\tilde{X} \in \R^{N \times D}$ is the perturbed version of $X$, and a matrix of binary values called the \textit{error mask}, $E \in \{0, 1\}^{N\times D}$, indicates the positions of errors in $\tilde{X}$.
If the value $\tilde{X}_{i,j}$ in row $i$ and column $j$ is missing, the entry $E_{i,j}$ is $1$, and $0$ otherwise.

\subsection{Learning the Error Dependencies}
\label{subsec:detecting-the-error-distribution}
To determine the error mechanism of a column $j$, \ourAlgorithm{} learns to predict the error mask of that column in three distinct supervised learning tasks:
The \textit{Complete} task uses all available information, the entire table serves as training data $X_{train} = X$, and the error mask of column $j$ serves as the target, $y = E_{:,j}$.
The \textit{Shuffled} task uses the entire table as training data $X_{train} = X$, and the randomly permuted error mask of column $j$ as the target, $y = \tilde{E}_{:,j}$.
The \textit{Excluded} task excludes column $j$ from the training data such that $X_{train} = X_{:,-j}$, and the error mask of column $j$ serves as the target, $y = E_{:,j}$.
These tasks are listed in~\autoref{tab:training-settings}.

\begin{table}[h]
\centering
\caption{Summary of the different training setups used by the MechDetect algorithm. 
}
\label{tab:training-settings}
\begin{tabular}{l l l}
\toprule
\textbf{Learning Task} & \textbf{Training Data} & \textbf{Target Data $y$} \\
\midrule
\textit{Complete} & $X$ & $E_{:,j}$ \\
\textit{Shuffled} & $X$ & $\tilde{E}_{:,j}$ \\
\textit{Excluded} & $X_{:,-j}$ & $E_{:,j}$ \\
\bottomrule
\end{tabular}
\end{table}

To perform the binary classification we utilize \texttt{HistGradient\-BoostingClassifier} from the scikit-learn~\cite{pedregosa2011scikit} library, which implements a tree-based method inspired by LightGBM~\cite{ke2017lightgbm}. We decide against using a linear model, as~\cite{wangScoreTestMissing2023} did, because we want \ourAlgorithm{} to support non-linear dependencies between the data and the error mask. Additionally, tree-based methods are robust and attain state-of-the-art performance on tabular datasets~\cite{grinsztajn2022tree}.
However, any supervised algorithm able to carry out a binary classification task is applicable.

We use 10-fold cross-validation to estimate a distribution of \textit{Area Under the Receiver Operating Characteristic Curve} (AUC-ROC) metrics~\cite{hanley1982meaning} to measure the model's ability to learn the target $y$ given the training data $X_{train}$ in the three tasks.
Each learning task results in a set of ten AUC-ROC measures $A = \{a_0, \dots, a_{9}\}$.

\subsection{The \ourAlgorithm{} Algorithm}
\label{subsec:algorithm}
\begin{figure}
    \centering
    \includegraphics{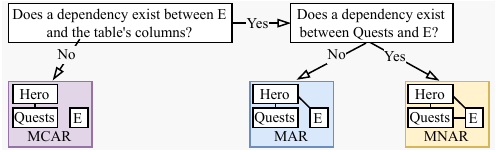}
    \caption{
    Example of applying \ourAlgorithm{} to the column Quests. The leaves representing the dependency structure for the corresponding error mechanism. For MCAR, there is no dependency between $E$ and the data. In the case of MAR mechanism, the error distribution potentially depends on the column Hero, whereas for MNAR, $E$ additionally depends on the column Quests.
    }
    \label{fig:testing-tree}
\end{figure}

An illustration of the two-step procedure that \ourAlgorithm{} uses to determine the error mechanism is shown in \autoref{fig:testing-tree}.
First, \ourAlgorithm{} performs a test to identify if the error mask $E_{:,j}$ depends on any data in table $X$.
If there is a dependency between the error mask $E_{:,j}$ and the data $X$, a second test is evaluated to determine if the column of interest, column $X_{:,j}$, specifically contains useful information about $E_{:,j}$.

\vspace{1em}
\noindent\textbf{Test 1: Do errors depend on data?}
If the classifier trained for the \textit{Complete} task obtains higher performance than the one trained for the \textit{Shuffled} task, there is a dependency between $X$ and $E_{:,j}$. This suggests the errors are either MAR or MNAR distributed.
Formally, the first two learning tasks result in two sets of ten AUC-ROC measures, called $A_{\text{Complete}}$ and $A_{\text{Shuffled}}$.
We compare the classifiers' performances on the tasks
by comparing the distributions of $A_{\text{Complete}}$ and $A_{\text{Shuffled}}$ through a Mann-Whitney-U-Test (MWU)~\cite{mann1947test} denoted $T_{\text{Shuffled}} = \text{MWU}(A_{\text{Complete}}, A_{\text{Shuffled}})$.
The null hypothesis of $T_{\text{Shuffled}}$ states that the distribution of $A_{\text{Complete}}$ is not stochastically larger than the distribution of $A_{\text{Shuffled}}$.
If the null hypothesis of $T_{\text{Shuffled}}$ is rejected, the evidence is consistent with the statement: the distribution of $A_{\text{Complete}}$ is stochastically larger than $A_{\text{Shuffled}}$. That is, the classifier from the \textit{Complete} task was able to learn a dependency between $X$ and the error mask $E_{:,j}$. This suggests that the errors are either MAR or MNAR. Conversely, if the null hypothesis cannot be rejected, we assume the errors to be MCAR. This is because there is insufficient statistical evidence to support the statement: $A_{\text{Complete}}$ is stochastically larger than $A_{\text{Shuffled}}$. We thereby conclude there is no dependency between the data distribution and the error distribution in column $j$.

\vspace{1em}
\noindent\textbf{Test 2: Do errors depend on values in column j?}
If the classifier trained on the \textit{Complete} task outperforms the one trained on the \textit{Excluded} task, this indicates that  $X_{:,j}$ contains useful information for predicting $E_{:,j}$, in turn suggesting that the missing values are MNAR.
Formally, we assess the difference in performances by comparing the distributions of $A_{\text{Complete}}$ and $A_{\text{Excluded}}$ through an additional Mann-Whitney-U-Test, denoted as $T_{\text{Excluded}} = \text{MWU}(A_{\text{Complete}}, A_{\text{Excluded}})$.
The null hypothesis of this test states that the distribution of $A_{\text{Complete}}$ is not stochastically larger than that of $A_{\text{Excluded}}$.
If the null hypothesis of $T_{\text{Excluded}}$ is rejected, the evidence supports the statement that the distribution of $A_{\text{Complete}}$ is stochastically larger than the distribution of $A_{\text{Excluded}}$, indicating that the data in column $X_{:,j}$ provides useful information for predicting the error mask $E_{:,j}$.
This suggests the errors are MNAR; otherwise we assume the errors are MAR, as the data in column $j$ is helpful for predicting the error position in column $j$.

Because we perform two tests in sequence, we correct for multiple comparisons by applying a
Bonferroni correction~\cite{shaffer1995multiple} to the significance levels of our tests.
Two limitations of this approach are worth emphasizing. For one, if the data in column $j$ are strongly correlated with data in other columns, this will also result in identical distributions of $T_{\text{Excluded}}$ and $T_{\text{Complete}}$. Second, we assume here that values in column $j$ are observed along with the error (here missingness) mask. This condition is met in our experiments, though in practice this assumption may not hold true.

\begin{algorithm}
\caption{\ourAlgorithm{}}
\label{alg:mechdetect}
\begin{algorithmic} 
\State \textbf{Input:} Tabular dataset $X$, error mask $E$, test column $j$, significance level $\alpha$
\State \textbf{Output:} Error mechanism $M$, $p$ values $(p_1, p_2)$
\For{each learning task $t \in \{\text{Complete}, \text{Shuffled}, \text{Excluded}\}$}
    \State Train binary classifier $\text{clf}_t$ on $k=10$ folds for task $t$.
    \State Obtain set $A_t$ of cross-validated AUC-ROC scores from $\text{clf}_t$.
\EndFor
\State Conduct test $T_{\text{Shuffled}}$ on $A_{\text{Complete}}$ and $A_{\text{Shuffled}}$ to get $p_1$
\State Conduct test $T_\text{Excluded}$ on $A_{\text{Complete}}$ and $A_{\text{Excluded}}$ to get $p_2$
\If {$p_1 < \frac{\alpha}{2}$}
    \If {$p_2 < \frac{\alpha}{2}$}
        \State $M =$ MNAR
    \Else
        \State $M =$ MAR
    \EndIf
\Else
    \State $M =$ MCAR, $p_2 =$ NA
\EndIf
\State \Return $M, (p_1, p_2)$
\end{algorithmic}
\end{algorithm}

\section{Experiments}
\label{sec:experiments}
For a comprehensive evaluation of \ourAlgorithm{} on real-world data, we use benchmarking datasets for supervised learning provided by Grinsztajn et al.~\cite{grinsztajn2022tree}, Fischer et al.~\cite{fischer2023openmlctr} and the OpenML benchmarking suites~\cite{bischl2021openmlbenchmarkingsuites} as well as datasets from data cleaning research in ~\cite{jagerDataImputationData2024}.
We select $101$ datasets in total, specifically those with no missing values beforehand, fewer than $100,000$ rows and $500$ columns. We make our implementation of \ourAlgorithm{}, together with all datasets, available in our fully reproducible code repository\footnote{\ifpeerreview \url{https://anonymous.4open.science/r/mechdetect-F505} \else \url{https://github.com/calgo-lab/MechDetect}\fi}.

Each dataset is perturbed by inserting missing values with different error generation mechanisms (MCAR, MNAR and MAR). We exclude the target column of the benchmark task from the experiments as we do not investigate the downstream predictive performance, but rather the error mechanisms on the data alone. For perturbations, we use the \textit{tab\_err} library~\cite{JungJaegerErrorModels2025}  
with error rates of 0.1, 0.25, 0.5, 0.75, 0.9.
For example, an error rate of $0.1$ means that an affected column comprising 100 values will contain $\lfloor 0.1 \cdot 100 \rfloor = 10$ errors after the perturbation procedure.
In this way, we obtain the error mask $E$, as well as the clean $X$ and perturbed $\tilde{X}$ versions of each of the $101$ datasets. 
Following Algorithm~\ref{alg:mechdetect}, we apply the \ourAlgorithm{} algorithm and examine its performance while varying the error mechanism with which $\tilde{X}$ is perturbed. For each introduced mechanism, we apply \ourAlgorithm{} $21920$ times.
In addition to the correctness of the detection, we record the classification performance for the three different learning tasks listed in~\autoref{tab:training-settings}, which we select from the error-free table $X$ in \autoref{subsec:accuracy-of-algorithm} and \autoref{subsec:auc-roc-analysis}. As expected, we observe worse results on the perturbed table, $\tilde{X}$ which are detailed in \autoref{subsec:perturbed-accuracy}.


\subsection{Effectiveness of \ourAlgorithm{}}
\label{subsec:accuracy-of-algorithm}

To test the effectiveness of \ourAlgorithm{}, we first generate perturbed datasets $\tilde{X}$ as described in \autoref{sec:experiments}.
We then run the algorithm and record its prediction.
Finally, we examine the algorithm's classification performance in inferring the mechanism used to inject errors in terms of accuracy, where accuracy is defined as:
\begin{equation*}
    \text{Accuracy} = \frac{TP + TN}{TP + TN + FP + FN}
\end{equation*}
\vspace{1mm}
\noindent with TP -- True Positives (\ourAlgorithm{} detected the same error distribution present in the data), TN -- True Negatives, FP -- False Positives, and FN -- False Negatives.

\begin{figure}[htbp]
    \centering
    \includegraphics{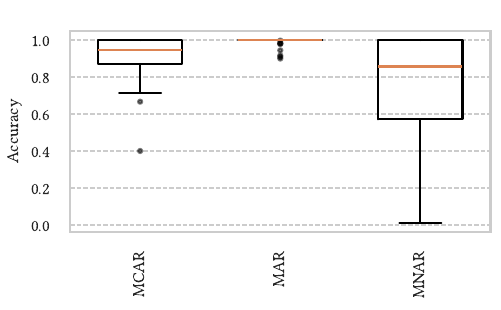}
    \caption{Accuracy of \ourAlgorithm{} classifying error mechanisms. The errors introduced are missing values, with an error rate of 0.5. We observe a mean accuracy of 89.04\% at this error rate.}
    \label{fig:mechdetect_accuracy_all_error_rates_clean}
\end{figure}

The observed accuracy scores for data perturbed with an error rate of 0.5 are visualized in~\autoref{fig:mechdetect_accuracy_all_error_rates_clean}.

\noindent The median accuracy across all datasets for MAR is $100$\%, meaning the tests detect columns that contain errors distributed at random almost perfectly.
Errors that are distributed completely at random (MCAR) are detected with a median accuracy of approximately $95$\%, exhibiting some dispersion. 
For errors distributed not at random (MNAR), median accuracy is about $86$\%, with measurements showing more dispersion.
As we explore in subsequent measurements, the characteristics of the classification are explained by the ability of the binary classifiers to approximate the error mask $E$ in the three learning tasks.
In summary, we observe a mean accuracy of $89.04$\% for determining the error mechanism in tables perturbed with an error rate of 0.5, clearly showing that \ourAlgorithm{} detects error mechanisms better than chance.
\begin{figure}[htbp]
    \centering
    \includegraphics{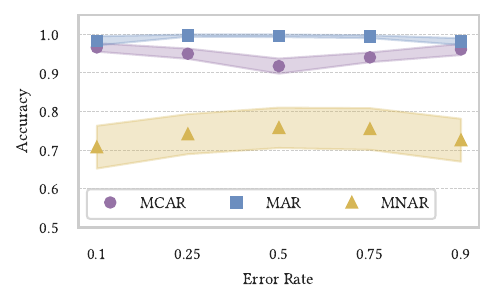}
    \caption{Mean accuracy of \ourAlgorithm{} as a function of the error rate. Colored areas around individual data points indicate the 95\% confidence interval for the mean.}
    \label{fig:mechdetect_accuracy_over_error_fraction}
\end{figure}

Furthermore, our measurements indicate that \ourAlgorithm{} infers the error mechanism at high accuracy for varying error rates.
In \autoref{fig:mechdetect_accuracy_over_error_fraction}, the mean accuracy for detecting the three error mechanisms is displayed as a function of the error rate.
We observe a decrease in the accuracy of \ourAlgorithm{} detecting MCAR errors as an error rate of $0.5$ is approached.
This decrease in accuracy in detecting MCAR errors is balanced by an increase in accuracy in detecting MNAR errors.
The high accuracy for detecting MAR errors appears largely unaffected by the error rate, as does the 95\% confidence interval for the mean.
In conclusion, we observe that \ourAlgorithm{} exhibits an accuracy for detecting the error mechanism that is largely robust for varying error rates, with a mean accuracy of $89.14$\% across all measured error rates and mechanisms.

\subsection{Classifier AUC-ROC Analysis}
\label{subsec:auc-roc-analysis}
\begin{figure}[h]
    \centering
\includegraphics{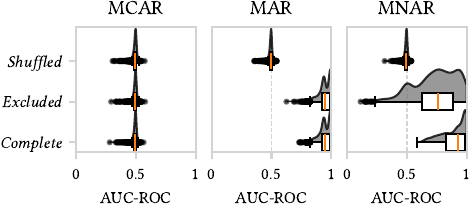}
    \caption{
    AUC-ROC scores of classifiers predicting the error mask from data. If errors are independent of the data (MCAR and \emph{Shuffled} setting), classifiers perform at chance level, as expected. If errors depend on data (MAR and MNAR), the \textit{Complete} and \textit{Excluded} classifiers achieve above chance performance, often close to AUC-ROC scores near 1.0. 
    }
    \label{fig:auc-roc-clean-0.5}
\end{figure}

In \autoref{fig:auc-roc-clean-0.5}, we investigate the performance of \ourAlgorithm{} further by examining the AUC-ROC scores of the binary classifiers from~\autoref{subsec:accuracy-of-algorithm} at an error rate of 0.5, the most challenging setting as we have observed empirically, for each supervised learning task (see~\autoref{subsec:detecting-the-error-distribution}).
The simplest case is MCAR, where there is no dependency between data and errors --- hence we expect the classifiers to perform as well as random guessing.
Indeed we find that the distribution of AUC-ROC scores is concentrated around 0.5 for MCAR in \autoref{fig:auc-roc-clean-0.5}. This indicates that the classifiers perform equally well, near random guessing performance.

In contrast, MAR and MNAR settings are characterized by a dependency between data and errors. This dependency is destroyed in the \emph{Shuffled} setting, and as expected the AUC-ROC scores in this case are around 0.5, at chance performance.
When the error distributions are dependent on the data however, the \textit{Excluded} and \textit{Complete} tasks have near identical distributions with centers close to 1.0, indicating that excluding column $j$ does not impact the classifier's performance substantially in the MAR setting.
One reason for this could be dependencies between the values in the respective column $j$ and values in other columns, as often observed in real world data sets.

In the MNAR case, omitting column $j$ from the training data in the \textit{Excluded} task reduces the AUC-ROC values substantially on average.
\ourAlgorithm{} detects this with the second test, and is therefore capable of distinguishing between MAR and MNAR error mechanisms.
If the values in column $j$ were independent from values in other columns, the distribution of AUC-ROC scores resulting from the \textit{Excluded} task would be clearly different from that resulting from the \textit{Complete} task, if the errors were MNAR.
However, as real-world datasets contain complex relationships between columns, information contained in column $j$ may be present in other columns as well, meaning that its absence would not reduce the classifier's performance as drastically as with stochastically independent columns.

\subsection{Training \ourAlgorithm{} on Perturbed Data}
\label{subsec:perturbed-accuracy}
\begin{figure}[h]
    \centering
    \includegraphics{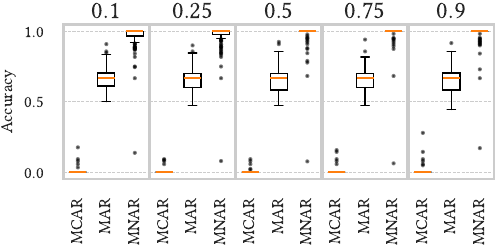}
    \caption{Accuracy of \ourAlgorithm{} using perturbed data $\tilde{X}$ to learn the tasks listed in \autoref{tab:training-settings}.
    Columns correspond to the error proportions in the masks.}
    \label{fig:mechdetect_accuracy_all_error_rates_dirty}
\end{figure}

To investigate how \ourAlgorithm{} performs on perturbed data, we perform additional experiments where no clean data is available.
In these experiments the data for the tasks (Training Data in~\autoref{tab:training-settings}) is sampled from the perturbed data $\tilde{X}$ instead of the clean data $X$.
Note that in the case of missing values, this implies that the missingness symbol is available to the classifiers used in the tests, except for the \textit{Excluded} task.
In other words, the classifier obtains a proxy for the target variable as an input variable.
This has two important implications for \ourAlgorithm{}: For one, the independence assumption of MCAR between data and errors is violated.
Second, that it is unlikely for the classifiers to learn any meaningful dependency between the data and the error, as the missingness symbol becomes a `shortcut' in the input data.
We emphasize that this limitation does not hold true for other types of errors apart from missingness.
The measurements from the perturbed data setting are displayed in~\autoref{fig:mechdetect_accuracy_all_error_rates_dirty}.
%

As expected, the results indicate that \ourAlgorithm{} does not detect the MCAR error mechanisms when applied to perturbed data. 
Since the error mask is trivially learned from the input data of the classifiers, the error position is not independent of the data.
The MCAR test detects this dependency and thus rejects the MCAR hypothesis, the median accuracy for MCAR is approximately $0.0$ with little dispersion. 
In contrast for the MNAR error distribution, the median accuracy is approximately $1.0$ with little dispersion, and the median accuracy for the MAR mechanism is approximately $0.65$ with moderate dispersion.

\begin{figure}[!htbp]
    \centering\includegraphics{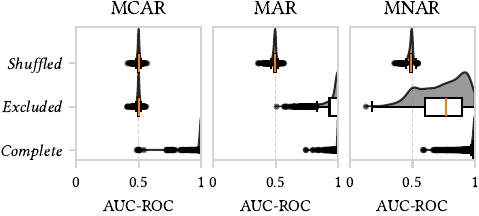}
    \caption{
    Distributions of AUC-ROC scores for detecting individual errors (here missing values) using perturbed data with missing value errors at an error rate of 0.5. The performance in the \textit{Complete} task is close to perfect as the classifiers observes the error signal in the input, which violates the independence assumption between data and errors of MCAR distributed erors.
    }
    \label{fig:auc-roc-dirty-0.5}
\end{figure}

In ~\autoref{fig:auc-roc-dirty-0.5}, we observe AUC-ROC performances for an error rate 0.5
that explain the error mechanism detection accuracy observed in~\autoref{fig:mechdetect_accuracy_all_error_rates_dirty}.
In the case of MCAR, the classifiers attain near perfect performance in the \textit{Complete} task on average, as the error position is encoded in the input variables. 
This leads \ourAlgorithm{} to falsely reject the MCAR case in favor of a wrong decision for MAR/MNAR for most datasets.
In the MAR case, we observe that the \textit{Excluded} and \textit{Complete} tasks exhibit different distributions of AUC-ROC scores, which also differ from the distributions observed of the same tasks on clean data in~\autoref{fig:auc-roc-clean-0.5}.
This difference in distributions explains the decreased performance for detecting MAR observed in~\autoref{fig:mechdetect_accuracy_all_error_rates_dirty}.
Finally in the MNAR case, the difference in performance distributions is captured well since the \textit{Complete} task has most values at $1.0$ while the \textit{Excluded} task deviates starkly, explaining the increased detection performance of MNAR errors.

\section{Conclusion}
\label{sec:conclusion}

The distribution of errors in a table is important information that researchers and data practitioners need to consider for correctly handling perturbed data in downstream tasks.
In this work, we introduced \ourAlgorithm{}, an algorithm to discern between the three error generation mechanisms MCAR, MAR and MNAR in tabular data.
We gave a formulation for the testing procedure underlying the algorithm in~\autoref{sec:methods} and performed an experimental analysis of the algorithm on 101 real-world datasets from four benchmarks in ~\autoref{sec:experiments}.
Our experiments show that \ourAlgorithm{} exhibits a mean accuracy of $89.14$\% in inferring the error mechanism across various error rates,
demonstrating that \ourAlgorithm{} is able to detect the error generating mechanism reliably when provided with the error mask and clean data. 

The availability of an error mask and clean training data is an important assumption \ourAlgorithm{} makes for detecting error dependencies --- an assumption driven by both practical applications and empirical results. 
From a practictioner's perspective, it is reasonable to assume that during development (though not necessarily during deployment) of ML components or data pipelines, clean data is available~\cite{schelter_challenges_2015}.
While cleaning methods can be used to obtain clean data from data perturbed by covariate shifts, missing data or other data quality problems~\cite{ilyasDataCleaning2019,xuchuHolisticDataCleaning2013}, empirical studies have shown that data driven error detection and cleaning requires clean data~\cite{jagerBenchmarkDataImputation2021}. 
In order to build data pipelines or ML components responsibly, some effort must be invested in obtaining clean data.
%
So while at first the clean data prerequisite might appear as a strong assumption, to the best of our knowledge, there is no prior work on detecting all three error mechanisms even under this assumption.
We thus consider our proposed approach a first step into the direction of detecting data dependent error mechanisms. 

In additional experiments we also investigated the performance \ourAlgorithm{} in scenarios when clean data is not available. In line with prior work on data cleaning\cite{jagerBenchmarkDataImputation2021} we find that without clean training data for calibrating \ourAlgorithm{} the detection of error mechanisms is impacted negatively. Specifically when \ourAlgorithm{} is calibrated on data perturbed with missing value errors generated with an MCAR mechanism, the mechanism cannot be reliably detected. This is expected as the errors are dependent on the data; after all the data contains the missingness indicator in the perturbed data. The assumption of independence between data and error mask in MCAR is hence violated and the MCAR hypothesis is falsely rejected.  
Future work could look to find techniques that allow the relaxation of the assumptions, specifically how to increase the performance of the algorithm when no clean data is available, and how the algorithm performs on perturbed data with errors other than missing values. 

\ifpeerreview
\else
\section{Acknowledgments}
\label{sec:acknowledgements}

This research was supported by the German Federal Ministry of Education
and Research grant number 16SV8856 and 16SV8835, by the Federal Ministry for Economic Affairs
and Climate Action of Germany for the RIWWER project
(Project number: 01MD22007H, 01MD22007C), by the Einstein Center
Digital Future, Berlin, and by the German Research Foundation (DFG) - Project
number: 528483508 - FIP 12.
\fi

\bibliographystyle{IEEEtran}
\bibliography{references}

\end{document}